\documentclass[conference]{IEEEtran}

\ifCLASSOPTIONcompsoc
  \usepackage[nocompress]{cite}
\else
  \usepackage{cite}
\fi

\ifCLASSINFOpdf
  \usepackage[pdftex]{graphicx}
\else
\fi

\usepackage{amsmath}
\usepackage{amssymb,amsfonts}

\usepackage{algorithm}
\usepackage{algorithmic}

\ifCLASSOPTIONcompsoc
    \usepackage[caption=false,font=footnotesize,labelfont=sf,textfont=sf]{subfig}
\else
 \usepackage[caption=false,font=footnotesize]{subfig}
\fi

\usepackage{url}

\usepackage{CJKutf8}

\usepackage{booktabs}
\usepackage{multirow}

\begin{document}
%
\title{Constructing Cross-lingual Consumer Health Vocabulary with Word-Embedding
from Comparable User Generated Content*\\
{\footnotesize \textsuperscript{*}Note: This manuscript has been accepted to the IEEE International Conference on Healthcare Informatics (IEEE ICHI 2024).}}


\author{\IEEEauthorblockN{Chia-Hsuan Chang}
\IEEEauthorblockA{\textit{College of Computing and Informatics} \\
\textit{Drexel University}\\
Philadelphia, USA \\
Email: shane.chang.tw@gmail.com}
\and
\IEEEauthorblockN{Lei Wang}
\IEEEauthorblockA{\textit{College of Computing and Informatics} \\
\textit{Drexel University}\\
Philadelphia, USA \\
Email: lw474@drexel.edu}
\and
\IEEEauthorblockN{Christopher C. Yang}
\IEEEauthorblockA{\textit{College of Computing and Informatics} \\
\textit{Drexel University}\\
Philadelphia, USA \\
Email: ccy24@drexel.edu}
}

\maketitle

\begin{abstract}
The online health community (OHC) is the primary channel for laypeople to share health information. To analyze the health consumer-generated content (HCGC) from the OHCs, identifying the colloquial medical expressions used by laypeople is a critical challenge. The open-access and collaborative consumer health vocabulary (OAC CHV) is the controlled vocabulary for addressing such a challenge. Nevertheless, OAC CHV is only available in English, limiting its applicability to other languages. This research proposes a cross-lingual automatic term recognition framework for extending the English CHV into a cross-lingual one. Our framework requires an English HCGC corpus and a non-English (i.e., Chinese in this study) HCGC corpus as inputs. Two monolingual word vector spaces are determined using the skip-gram algorithm so that each space encodes common word associations from laypeople within a language. Based on the isometry assumption, the framework aligns two monolingual spaces into a bilingual word vector space, where we employ cosine similarity as a metric for identifying semantically similar words across languages. The experimental results demonstrate that our framework outperforms the other two large language models in identifying CHV across languages. Our framework only requires raw HCGC corpora and a limited size of medical translations, reducing human efforts in compiling cross-lingual CHV.\end{abstract}


%
\IEEEpeerreviewmaketitle

\begin{CJK*}{UTF8}{bsmi}
\section{Introduction}

The increasing popularity of social media has facilitated discussion between Internet users and encourages us to share information and experiences. Such development contributes to the emergence of online health communities (OHCs). It stimulates a new channel for exchanging healthcare knowledge, including personal experiences of diseases, reflections on medicines, and communications between physicians and patients \cite{DeAndrea2016HowMedia,Zhou2018HarnessingManagement}. OHCs possess a wide variety of health consumer-generated content (HCGC). Analyzing HCGC helps study a wide range of research questions, such as the doctors-patients interaction \cite{Zhang2020-eo,Chen2020-na,Khurana2019-me} and health information search behavior \cite{Wu2019-es}. HCGC also serves as a valuable resource for extracting consumer health vocabulary~\cite{Zeng2007TermDevelopment,Doing-Harris2011Computer-AssistedData,Jiang2013UsingData,Vydiswaran2014MiningText,He2017EnrichingApproach,houMiningStandardizingChinese2018,Gu2019DevelopmentApproach,ibrahimAutomatedMethodEnrich2021}, identifying authoritative health answers \cite{Deardorff2017-av}, and facilitating many consumer-oriented applications such as language simplification system~\cite{moramarcoMorePatientFriendly2022a}.

However, analyzing HCGC is challenging because the vocabulary used by consumers is very different from that used in the medical literature and electronic health records. For example, consumers tend to use colloquial expressions like \textit{watery stool} rather than professional jargon such as \textit{diarrhea} for describing their bowel suffering. The open-access and collaborative consumer health vocabulary (OAC CHV)~\cite{Zeng2006ExploringVocabularies,Zeng2007TermDevelopment} is compiled to address this language gap. Compared with other professional-oriented vocabularies organized in the Metathesaurus of Unified Medical Language System (UMLS), OAC CHV intends to capture the medical expressions used by consumers, who are usually laypeople.

\subsection{Motivations}

We have two motivations for driving this study. The first is the underdevelopment of automatic term recognition (ATR) methods for identifying CHV across languages. In order to recognize the expression evolution and enrich the original OAC CHV \cite{Zeng2006ExploringVocabularies}, several studies propose various monolingual ATR methods such as applying string patterns \cite{Zeng2007TermDevelopment,Zeng2012SynonymDocuments,Vydiswaran2014MiningText,houMiningStandardizingChinese2018}, co-occurrence analysis \cite{Jiang2013UsingData}, machine learning models \cite{Zeng2007TermDevelopment, He2017EnrichingApproach}, and word embedding techniques \cite{Gu2019DevelopmentApproach,ibrahimAutomatedMethodEnrich2021}. Unfortunately, most studies expand English vocabulary and rely on language-specific features that barely apply to other languages. Besides, most cross-lingual ATR methods rely on costly resources, such as parallel corpus~\cite{Lu2008UsingRetrieval} and manual annotations~\cite{chapmanExtendingNegExLexicon2013a,Alfano2018-kt,rahimiWikiUMLSAligningUMLS2020a,Chen2020MultilingualizationApproaches,wajsburtMedicalConceptNormalization2021}, which are labor-intensive to build up.

The second motivation is that the lack of cross-lingual CHV prevents the development of consumer-oriented healthcare applications for non-English languages. Although English dominates the health information on the Internet, only 379 million people, which represent 4.9\% of the world population, are English native speakers. Considering the initiative proposed by the World Health Organization, ``Bridging the Language Divide in Health\footnote{\url{http://dx.doi.org/10.2471/BLT.15.020615}}'', there is still a large population of laypeople using their native languages to share and search for health information. Nowadays, language is a primary barrier for laypeople to search relevant health information on the Internet \cite{Teixeira_Lopes2017-kt,changBridgingConsumerHealth2023}.

\subsection{Research Objectives \& Contributions}

This study aims at proposing a cross-lingual ATR framework that helps extend the English CHV into other languages. The framework starts with collecting HCGC corpora of two different languages. With the corpora, the framework adopts word2vec techniques \cite{Mikolov2013DistributedCompositionality} on each monolingual corpus to learn the word associations used by laypeople. To identify similar medical concepts across languages, we align monolingual word spaces into a cross-lingual word space \cite{Smith2017-rj,lample2018word} using a small size of medical concept translations. The contributions of this study are as follows.
\begin{enumerate}
    \item Compared with previous cross-lingual ATR frameworks, we propose a resource-efficient framework to identify CHV across languages, which only requires a small set of medical entity translations and the HCGC corpus of each language.
    \item Compared with advanced large language models (LLMs), the cross-lingual word space induced by our framework outperforms two state-of-the-art LLMs (i.e., GPT-3.5-Turbo and Cohere Rerank) in identifying CHV across languages, showing an economic benefit of using our framework.
    \item We design a retrieval mechanism, dynamic threshold, to retrieve more quality CHV translations based on our cross-lingual word space.
\end{enumerate}

The rest of the paper is structured as follows: Section~II reviews related works in monolingual ATR methods and cross-lingual ATR methods. Section~III illustrates our proposed cross-lingual ATR framework, and Section~IV describes the experimental settings, metrics, and results. We then discuss implications, limitations, and future directions in Section~V. Finally, we draw a conclusion in Section~VI.

\section{Literature Review}

We review previous ATR techniques for expanding the medical vocabulary and categorize them into Monolingual ATR and Cross-lingual ATR. In the following, we discuss the limitations and inspirations of these studies.

\subsection{Monolingual ATR}

Previous studies on monolingual ATR relied on lexical patterns. Zeng et al. \cite{Zeng2007TermDevelopment} utilized MedlinePlus query logs for ATR, employing three components: identifying string candidates, expert collaboration for noise reduction, and estimating each remaining candidate's score (i.e., termhood) using nested string patterns \cite{Frantzi2000AutomaticMethod}. Doing-Harris \& Zeng-Treitler \cite{Doing-Harris2011Computer-AssistedData} adopted a similar approach on the collected HCGC from PatientLikeMe.com for ATR. Vydiswaran et al. \cite{Vydiswaran2014MiningText} proposed string rule mapping (e.g., also called, also known as, and also referred to as) to extract synonyms from health-related Wikipedia pages. Likewise, Hou et al. \cite{houMiningStandardizingChinese2018} used string patterns, e.g., 又稱\text{ }(as known as) to identify consumer-used disease terms from a Chinese online health forum. Instead of using ad-hoc lexical patterns, Jiang and Yang \cite{Jiang2013UsingData} modified the \textit{tf-idf} technique and applied it on MedHelp threads for extracting frequent co-occurring lay terms. He et al. \cite{He2017EnrichingApproach} used clustering with combined lexical, syntactic, and contextual features and extracted potential term candidates from \textit{Yahoo Know!} Q\&As. Gu et al. \cite{Gu2019DevelopmentApproach} and Ibrahim et al. \cite{ibrahimAutomatedMethodEnrich2021} proposed word embedding techniques and identified semantically-similar terms from collected health corpora. 

Most previous monolingual ATR methods required string patterns and lexical features that were only applicable to English, which causes difficulty in generalizing to other languages. Even though co-occurrence calculation and word embedding determination are language-agnostic techniques, a clearer way to apply them across languages is needed.

\subsection{Cross-lingual ATR}

Marko et al. \cite{Marko2006TowardsLexicon} and Campillos-Llanos \cite{campillos-llanosFirstStepsBuilding2019a} depended on morpho-syntactic features for linking potential translation and synonyms across languages. However, the morphological pattern only worked for cognate languages, such as Indo-European languages, that shared linguistic connections. Lu et al. \cite{Lu2008UsingRetrieval} created a parallel corpus, which utilized both Web anchor texts and search result pages. They then used the parallel corpus to identify high-quality English--Chinese medical term pairs. On the other hand, Chapman et al. \cite{chapmanExtendingNegExLexicon2013a} and Alfano et al. \cite{Alfano2018-kt} recruited a group of translators to translate English CHV into other languages. Rahimi et al. \cite{rahimiWikiUMLSAligningUMLS2020a} relied on aliases across languages on the Wikidata to link UMLS concepts. Both parallel corpus preparation and manual translation process took much work to scale and had coverage concerns.

As morphological patterns hardly worked for non-cognate languages. Moreover, previous cross-lingual ATR methods relied on expensive resources, such as parallel corpus and many manual annotations. In this study, we leveraged the space alignment technique \cite{Smith2017-rj} for aligning the word contextual information across languages, which was applicable across any language pairs \cite{lample2018word} and only required very few manual supervisions.

\section{The Cross-lingual ATR Framework}

\begin{figure}[t]
\centering
\includegraphics[width=\columnwidth]{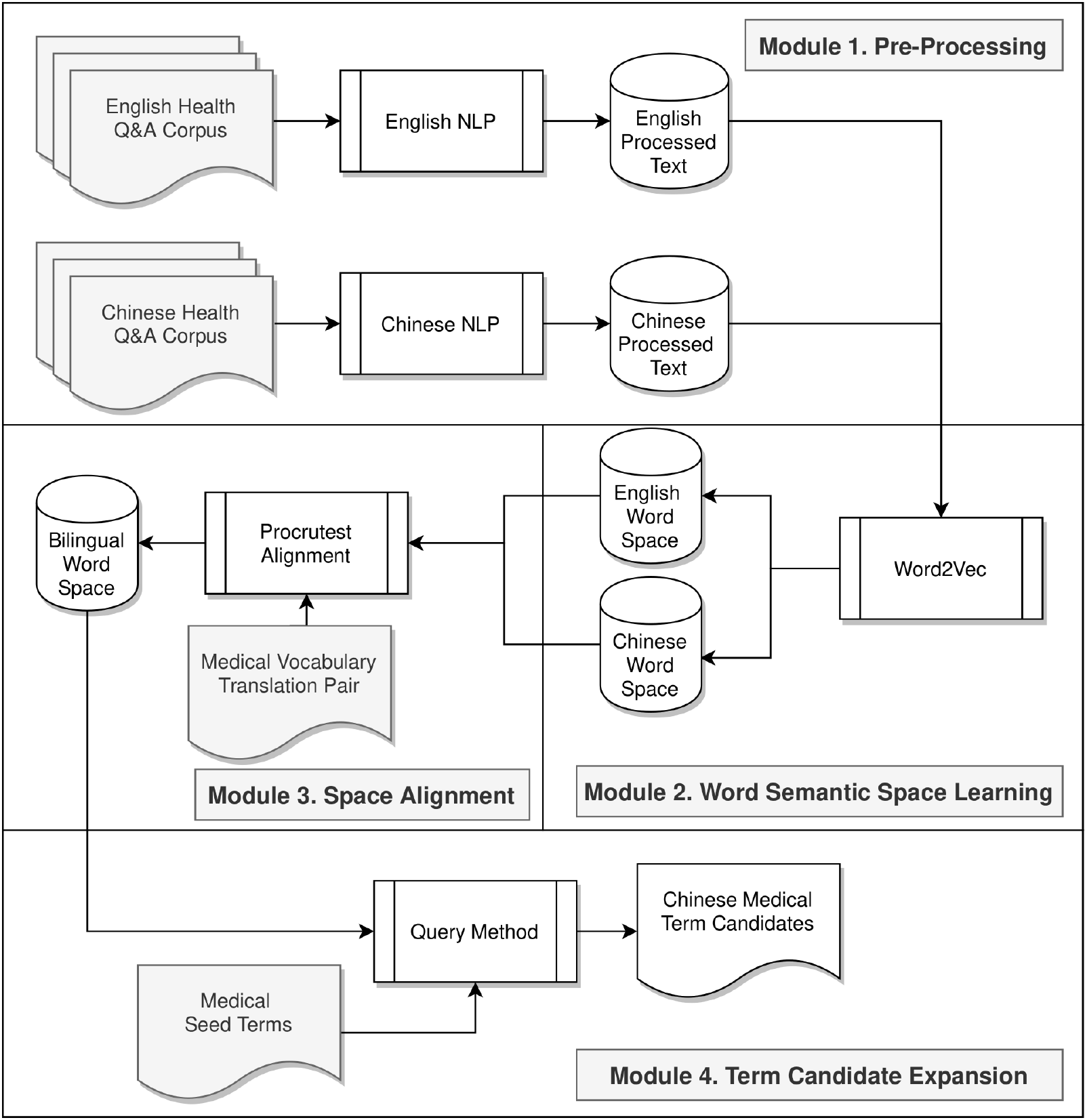}
\caption{Our framework for bilingual medical term expansion}
\label{fig: the procedure of methodology}
\end{figure}

Fig. \ref{fig: the procedure of methodology} illustrates our cross-lingual ATR framework for expanding CHV across languages. Our framework consists of four main modules. With collected health Q\&A corpora, the first \textit{pre-processing module} employs existing NLP toolkits to clean the raw texts and transforms them into processed contents. Subsequently, using a word embedding technique, the \textit{word semantic space learning module} determines the word vector space of each language from collected corpora. To compare words in a language-agnostic way, the \textit{space alignment module} pairs up the spaces across languages utilizing a set of bilingual medical translation pairs as anchors in the spaces. Given the bilingual word space, the last \textit{term candidate expansion module} requires a set of seed words, e.g., OAC CHV \cite{Zeng2006ExploringVocabularies}, to expand their synonyms for eliciting the bilingual CHV. In the following subsections, we present the data collection and detail each module of our cross-lingual ATR framework.


\subsection{Data Collection}

We collected the healthcare Q\&A corpus from OHCs, the platforms for laypeople to exchange health-related information, where people tend to use colloquial expressions rather than professional jargon. Hence, we can exploit the usage of health vocabulary from such HCGCs on the OHCs.

MedHelp is a popular English OHC consisting of various communities for consumers to discuss diseases. We extracted 520,659 discussion threads from 106 different communities in MedHelp. To gather Chinese contents, we collected three Chinese healthcare Q\&A websites, including eDoctor\footnote{\url{http://taiwanedoctor.mohw.gov.tw/}}, MED-NET\footnote{\url{https://expert.med-net.com/index\#/index}}, and Kingnet\footnote{\url{https://www.kingnet.com.tw/knNew/inquiry/outpatient.html}}, to retrieve 259,709 inquires and their responses, where most questions are answered by doctors and solved by one response. Table \ref{tab:dataset-descriptive-statistics} presents descriptive statistics of two corpora, in which the English corpus has more textual content (meaured by characters) than the Chinese one.

\begin{table}[t]
\centering
\caption{Descriptive statistics of our collected corpora.}
\label{tab:dataset-descriptive-statistics}
\resizebox{\columnwidth}{!}{%
\begin{tabular}{@{}lllll@{}}
\toprule
               & No. of Documents  & No. of Sentences & Avg. Content Length \\ \midrule
English Corpus & 520,659 & 15.82M      & 754.11 \\
Chinese Corpus & 259,709 & 0.70M       & 177.70 \\ \bottomrule
\end{tabular}
}
\end{table}

\subsection{Pre-processing}

After collecting the corpora, we processed our textual data using existing NLP toolkits. For English, we applied sciSpacy \cite{Neumann2019ScispaCy:Processing}, a package for biomedical text processing, for word tokenization and medical concepts extraction by consulting Metathesaurus of UMLS \cite{Bodenreider2004TheTerminology}. Regarding pre-processing Chinese texts, we employed jieba\footnote{\url{https://github.com/fxsjy/jieba}} for word tokenization. Due to the lack of a reliable medical name entity recognition tool in the Chinese language, we collected 7,824 Chinese medical entities from the healthcare category of BaiduBaike\footnote{\url{https://baike.baidu.com/wikitag/taglist?tagId=76625}}, and 11,604 Chinese medical entities from Wikipedia. The two Chinese entity sets were combined as our custom dictionary for entity extraction. To prevent breaking multiword expressions with high co-occurrences into separate words, we utilized the phrases identification method \cite{Mikolov2013DistributedCompositionality} to connect them. We filtered out the noises, including stop words, special symbols, and punctuations for both languages.

\subsection{Determining Monolingual Word Vector Space}

In the second module, we leverage the word embedding technique to determine the word vector space of each language. The word vector space helps us compare semantic similarities between words. One of the powerful word embedding algorithms is Skip-gram \cite{Mikolov2013DistributedCompositionality}, known for preserving the word semantics and capable of encoding each word into a continuous vector representation. By utilizing Skip-gram, we determined two word vector spaces $V^{en} \in \mathbb{R}^{|W^{en}| \times d}$ and $V^{zh} \in \mathbb{R}^{|W^{zh}| \times d}$ for English and Chinese healthcare Q\&A corpus, respectively. Note that $W^{en} (W^{zh})$ is the set of words appearing in the English (Chinese) corpus, and $d$ is the size of hidden dimensions representing each word. Moreover, we normalize every word vector by applying $l2$-norm, where $V^l_i$ is a vector of word $i$ in language $l$. With the normalization, we obtain two $d$-dimensional spheres, each for one language.


\subsection{Aligning the Spaces of Different Languages}

With the contextual information of words determined from the word semantic space learning module, the word spaces were determined independently, resulting in unaligned word dimensions between languages. We aligned consumer health vocabularies of different languages in the space alignment module. To facilitate the alignment procedure, we assumed the space structures were topologically similar, also called isometric in previous studies, across languages \cite{Smith2017-rj,lample2018word}. This assumption allowed us to use a linear transformation matrix $L \in \mathbb{R}^{d \times d}$ for aligning one source word space into one target word space. In our setting, we treated the Chinese word space $V^{zh}$ as source word space and applied $L$ on it to align with the target word space -- English word space $V^{en}$. 

Determining $L$ required a set of translation pairs $P \subseteq W^{en} \times W^{zh}$ as anchor points between the two word spaces. We leveraged Wikipedia as our knowledge source to identify quality $P$ for our two health-oriented word spaces. Because of collective intelligence and multilingualism, Wikipedia possesses numerous health entities compiled into several languages. The title of each page is usually the name of an entity. Hence, we followed the data collection strategy proposed in \cite{Vydiswaran2014MiningText} to collect a bilingual title list of health-related pages from Wikipedia. In detail, we adopted the MediaWiki Action API to traverse the category hierarchy organized in Wikipedia and drill down subcategories for \textit{health} category with a depth of five layers. We then recorded the English title of each relevant page and its Chinese title as one translation pair by querying the \textit{langlink} property in MediaWiki Action API. For instance, using ``Diarrhea'' as a query, we find its corresponding Chinese title ``腹瀉'' (Diarrhea)~\footnote{The retrieved result can be verified by using \url{https://en.wikipedia.org/w/api.php?action=query&titles=Diarrhea&prop=langlinks&lllang=zh&format=json}.}.


After compiling $P$, to optimize $L$, each title pair were represented as a vector pair by consulting $V^{en}$ and $V^{zh}$, which resulted into two subspaces: $V^{en}_{\dagger} \in \mathbb{R}^{|P| \times d}$ and $V^{zh}_{\dagger} \in \mathbb{R}^{|P| \times d}$. The objective of $L$ aims at minimizing the following equation:

\begin{equation} \label{eq: orthogonal projection}
    \mathop{\arg\min}_{L} \| {V^{zh}_{\dagger}}L - V^{en}_{\dagger} \|^{F} \text{ s.t. } L^TL=I,
\end{equation}

\noindent where the orthogonal constraint of $L$ keeps the density between words in $V^{zh}_{\dagger}$ invariant after transforming. Solving $L$ is hereby an orthogonal Procrustes problem which has a analytical solution \cite{lample2018word} using singular value decomposition (SVD): 
\begin{equation} \label{eq: proscrutes alignment}
    L = UV, \text{with }U{\Sigma}V^T = SVD(({V^{zh}_{\dagger}})^TV^{en}_{\dagger}).
\end{equation}

\begin{figure*}[t]
\centering
\subfloat[$V^{zh}$]{
\includegraphics[width=.3\textwidth]{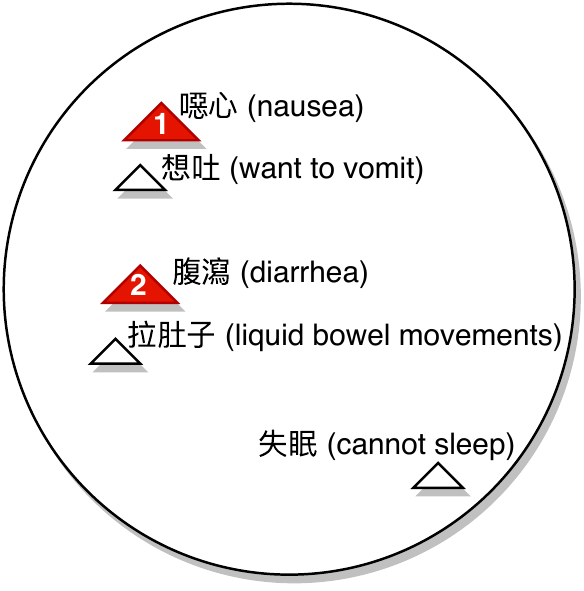} \label{subfig. Chinese word space}}
\hfill
\subfloat[$V^{en}$]{
\includegraphics[width=.3\textwidth]{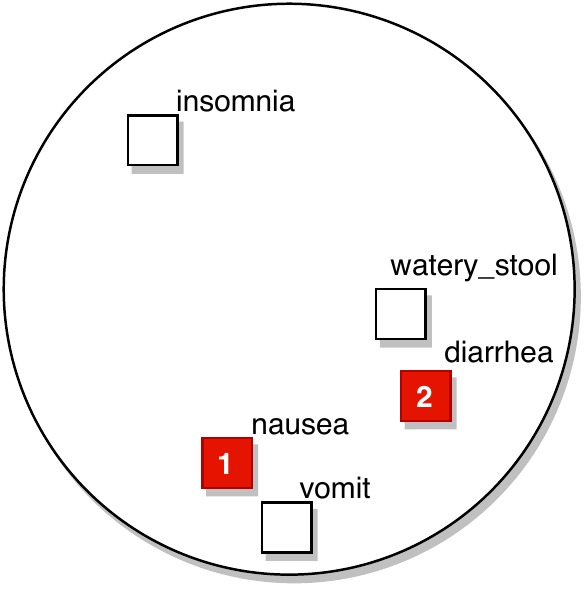}
\label{subfig. English word space}}
\hfill
\subfloat[$V^{bi}$]{
\includegraphics[width=.3\textwidth]{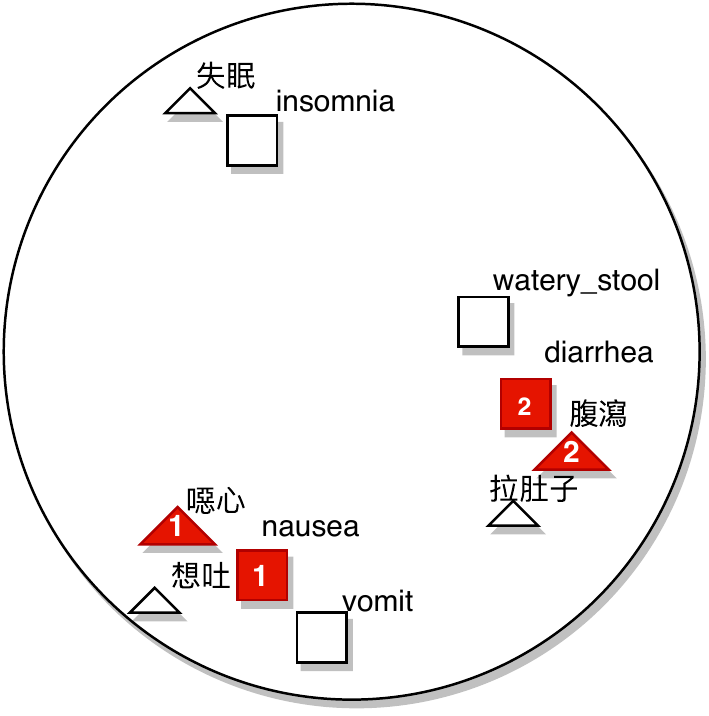}
\label{subfig. bilingual word space}}
\caption{The illustration of the spaces alignment}
\label{fig: the space illustration}
\end{figure*}

To better illustrate the alignment procedure, we provide an example in Fig. \ref{fig: the space illustration}, where the red items with the same number in Fig. \ref{subfig. Chinese word space} and \ref{subfig. English word space} represent anchor pairs in the $P$. The goal of $L$ is to transform $V^{zh}$ to align each anchor pair using the $V^{en}$ as a standard reference. When $L$ is determined, the bilingual word space is derived as Eq. \ref{eq. bilingual word space}.

\begin{equation} \label{eq. bilingual word space}
    V^{bi} = (V^{zh} \cdot L) \cup V^{en},
\end{equation}

\noindent where $V^{bi} \in \mathbb{R}^{|W^{bi}| \times d}$ and $W^{bi} = W^{en}+W^{zh}$.

The alignment module not only aligned anchor word pairs in $P$ (red items in Fig. \ref{fig: the space illustration}) but also indirectly aligned non-anchor words (i.e., those words are not in $P$) across languages. Fig. \ref{subfig. bilingual word space} explains three possible alignment cases between non-anchor words across languages.

\begin{enumerate}
    \item \textbf{Case 1: Non-anchor words close to the anchors.} Even though only anchors involve in determining $L$, we could expect to find more synonyms after transformation due to the similar representations encoded in neighbors of anchors. For example, given diarrhea and nausea as queries, we not only can find their corresponding anchors 腹瀉\text{ }(diarrhea) and 噁心\text{ }(nausea), but also 拉肚子\text{ }(liquid bowel movements) as well as 想吐\text{ }(want to vomit), respectively.
    \item \textbf{Case 2: Non-anchor words distributed similarly across languages.} As we transform the whole source space into the target space, the words with similar meanings will match up by $L$ due to the assumption of topological similarity between spaces. For example, even if we do not provide \textit{insomnia} or its neighborhoods as anchors, we still could find its Chinese synonym 失眠
   \text{ }(cannot sleep). This is because similar word concepts are distributed closely across languages after aligning.
    \item \textbf{Case 3: Non-anchor words that bias at a particular language space.} As shown in Table \ref{tab:dataset-descriptive-statistics}, the size of corpus is significantly different between languages. Besides, some diseases, drugs, or other medical entities may differ between languages, which causes some areas to be monolingual in the resultant bilingual word space. In this situation, given an English word, its Chinese neighbors have significantly greater distances than the ones in the previous two cases.
\end{enumerate}

\subsection{Word Expansion by Querying the Spaces}

Given the bilingual word space, the words should be language-agnostic; hence, we can compute semantic closeness between any two words using the cosine similarity: $\text{cos}(V^{bi}_i,V^{bi}_j)$, where $i, j \in W^{bi}$ and $V^{bi}_i$ is the word vector of word $i$. Given a set of seed words $Q$ of a language as queries, we can search for their synonym candidates across languages using the bilingual word space. In this study, seed words could be either English or Chinese, so we have two query directions: (1) EN $\rightarrow$ ZH, and (2) ZH $\rightarrow$ EN. For each seed word $q \in Q$, we find the nearest words as candidates. To ensure the expansion quality, we only kept the candidate $r$ whose similarity with the query beyond a threshold ($\delta$): $\text{cos}(q, r) >= \delta$.





\textbf{Dynamic Threshold.} Instead of using only one $\delta$ to filter out noisy synonym candidates, we also consider using different thresholds for different queries. This is because we observe that it is not easy to find an appropriate $\delta$ for all queries. Take Table \ref{tab:example of two queries with different modularity} as an example, we could set $0.64$ as $\delta$ to filter out the noisy candidates of ``高血壓'' (hypertension), but this $\delta$ is too strict for ``雞眼'' (wart). On the other hand, if we lower the $\delta$ for ``雞眼'' (wart), the $\delta$ is adversely too vague for ``高血壓'' (hypertension). The three cases can explain this phenomenon resulting from the space alignment module. Some queries are distributed equally across languages because they belong to case 1 or 2. In contrast, some queries are biased toward a language that makes them far away from their synonyms of the other language. 

\begin{table}[t]
\caption{The results of two queries: \\``高血壓'' (hypertension) and ``雞眼'' (wart)}
\label{tab:example of two queries with different modularity}
    \resizebox{\columnwidth}{!}{%
    \begin{tabular}{@{}llllll@{}}
    \toprule
    Query & Candidate & $cos(q,r)$ & Query & Candidate & $cos(q,r)$ \\ \midrule
    \multirow{10}{*}{高血壓} & hypertension & 0.81 & \multirow{10}{*}{雞眼} & wart & 0.58 \\
     & htn & 0.74 &  & plantar\_wart & 0.55 \\
     & high\_bp & 0.73 &  & callus & 0.54 \\
     & hbp & 0.72 &  & splinter & 0.54 \\
     & high\_blood\_pressure & 0.72 &  & ganglion\_cyst & 0.53 \\
     & hypertention & 0.65 &  & skin\_tag & 0.52 \\
     & hypertensive & 0.64 &  & ingrown\_toenail & 0.52 \\
     & high\_cholesterol & 0.62 &  & bunion & 0.51 \\
     & hyperlipidemia & 0.61 &  & athlete\_foot & 0.51 \\
     & blood\_pressure\_problem & 0.61 &  & hard\_skin & 0.50 \\ \bottomrule
    \end{tabular}%
    }
\end{table}

Our dynamic threshold mechanism is based on modularity. Modularity is a metric to evaluate the bias level of a query in a specific language in the bilingual word vector space. The lower modularity of a query means more words of the other language surrounding that query. The higher modularity of a query instead means that the query has more words of the same language as its neighbors. To measure the modularity, we refer to the previous work \cite{Fujinuma2019AModularity} and design our calculation for approximating the modularity score in the Algorithm \ref{algo: The algorithm for calculating modularity}. Given a query $q$, \textit{Line 1} and \textit{Line 2} collect top-$K$ English and Chinese nearest neighbors as two sets of words $G^{en}$ and $G^{zh}$, respectively. \textit{Line 3} and \textit{Line 4} are then used to calculate an average score of cosine similarity $\eta^{en}(\eta^{zh})$ for $G^{en}(G^{zh})$. As \textit{Line 5}, the modularity is the difference between $\eta^{en}$ and $\eta^{zh}$. Therefore, the lower modularity of a query indicates that its nearest neighbors of the other language are close. In other words, when a query with lower modularity, the more possibilities the query is either in case 1 or case 2. Otherwise, when a query with higher modularity, it will possibly be case 3.

\begin{algorithm}
\caption{The algorithm for calculating modularity} \label{algo: The algorithm for calculating modularity}
    \begin{algorithmic}[1]
        \REQUIRE bilingual word space $V^{bi}$, a query $q$, number of neighbors $K$
        \STATE $G^{en}$ = top-$K$ similar English words of $q$
        \STATE $G^{zh}$ = top-$K$ similar Chinese words of $q$
        \STATE $\eta^{en} = \mathbb{E}_{r \sim G^{en}}[\text{cos}(q,r)]$
        \STATE $\eta^{zh} = \mathbb{E}_{r \sim G^{zh}}[\text{cos}(q,r)]$
        \RETURN $m = |\eta^{en} - \eta^{zh}|$
    \end{algorithmic}
\end{algorithm}

To dynamically set different $\delta$ for queries, we applied Algorithm \ref{algo: The algorithm for calculating modularity} on each seed word $q$. After sorting the seed words ascendingly with respect to their modularity scores, we grouped $Q$ into $N$ subsets of equal size based on quantiles $\{Q_i\}^{N}_{i=1}$, in which $Q_1$ contains the queries with the lowest modularity. As a result, we searched the best $\delta$ of each subset with regard to a determined metric (e.g., F1-score).

\section{Evaluation}

\subsection{Experiment Setups}

\textbf{Parameter Settings.} We determined the word vector space of each language by feeding all texts from English and Chinese healthcare Q\&A corpus. We employed the word2vec module from \textit{gensim}\footnote{\url{https://radimrehurek.com/gensim/models/word2vec.html}}, which implements the Skip-gram algorithm \cite{Mikolov2013DistributedCompositionality}. All parameters of the Skip-gram were default values. The size of vocabulary for English and Chinese word space are 54,061 and 17,882, respectively. We used 719 bilingual title pairs from Wikipedia to enable the process of space alignment. Note that we collected 11,604 pairs in total, yet we dropped a pair if either English or Chinese title in a pair did not exist in our word spaces. In our further evaluations, we demonstrated effective performances even with a small proportion of words as anchors.


\noindent \textbf{Test Set Preparation.} We evaluated our proposed framework by testing the query performance of frequently used medical entities selected from our collected healthcare Q\&A corpora. Only five divisions (community groups in MedHelp), including General-Health, Women-Health, Dermatology, Ear-Nose-Throat, and Neurology, were involved in the further evaluations because these divisions were the most popular topics in our corpora. To generate English test queries, we calculated the frequency of each term that was selected from the OAC CHV \cite{Zeng2006ExploringVocabularies} and only retained terms that belonged to two semantic types: \textit{Sign or Symptom} [T184] and \textit{Disease or Syndrome} [T047]\footnote{Both T184 and T047 were the semantic types defined in the Semantic Network of UMLS.}. After filtering out the general terms (i.e., terms appearing in all divisions), we picked the top 10 frequent words as queries from each division, resulting in 50 English queries. On the other hand, we sorted every word by its frequency within each division in the Chinese Q\&A corpus to select Chinese queries. We filtered out the words that did not represent a sign, symptom, disease, or syndrome. After that, we picked each division's top 10 frequent words as queries from the remaining words. As a result, we had 50 Chinese queries. Subsequently, we manually scrutinized the top 100 nearest neighbors of each Chinese and English query to make the ground truth. The labeling strategy was as follows: If a neighbor word has the same semantic meaning as the query, it will be labeled as a relevant item, otherwise a non-relevant item. Three annotators were responsible for the labeling procedure. All annotators had research experiences in health informatics and are Chinese-English bilingual speakers. The inter-rater agreement was measured using Cohen's kappa coefficient ($\kappa$) and reached $0.83$. All conflict annotations were resolved by majority rule.

\noindent \textbf{Two Experiments.} We conducted two experiments and reported the results in the following subsections. The first experiment compared the quality of bilingual health word embedding induced by our proposed framework with other existing language models for identifying CHV across languages. The second experiment evaluated the sensitivity of the parameters of our proposed framework, mainly focused on parameters for retrieval, including similarity threshold~$\delta$ and dynamic threshold mechanism.

\subsection{Experiment 1: Model Comparison}

\textbf{Performance Metrics.} We used mean reciprocal rank (MRR) as the metric for evaluating the retrieval performances of different systems. For each query~$q_i$, we identified the rank of its first relevant result~$rank_i$. As such, given a set of queries and ranks~$Qr=\{q_i: rank_i, 1 \leq i \leq |Qr|\}$, MRR is the average of reciprocal ranks for~$Qr$, calculated as follows:

\begin{equation}
    MRR = \frac{1}{|Qr|}\sum^{|Qr|}_{i=1}\frac{1}{rank_i}
\end{equation}

When the system places the relevant item in the first place, its MRR will be $1$. In our case, high MRR means relevant translations are identified earlier, which prevents people from reading through many irrelevant translations. Reversely, if a system hardly identifies relevant items or places those items in the latter positions, the MRR will be close to 0.

\noindent \textbf{Competitive Models.} We compared our induced bilingual health word embedding with (1) OpenAI's GPT-3.5-Turbo and (2) Cohere's Rerank. Both models were easily accessed by their official APIs and were the most common and accessible resources for most NLP tasks.

\begin{enumerate}
    \item GPT-3.5-Turbo~\footnote{We have considered BioGPT, health-specific GPT; however, it is pre-trained only using English corpora.}, it is pre-trained using a significant amount of web data; it should learn about health vocabulary usage from the online health communities of different languages. It also performs well on various NLP tasks, including translation task~\cite{hendyHowGoodAre2023}.
    \item Rerank~\footnote{We have considered OpenAI's embedding API, but the official document does not suggest the usage beyond English.}, it is supported by rerank-multilingual-v2.0, an LLM pre-trained using a set of multilingual corpora. Its ability on cross-lingual information retrieval has been demonstrated~\cite{kamallooEvaluatingEmbeddingAPIs2023}.
\end{enumerate}

\begin{figure*}[t]
\centering
\includegraphics[width=0.85\textwidth]{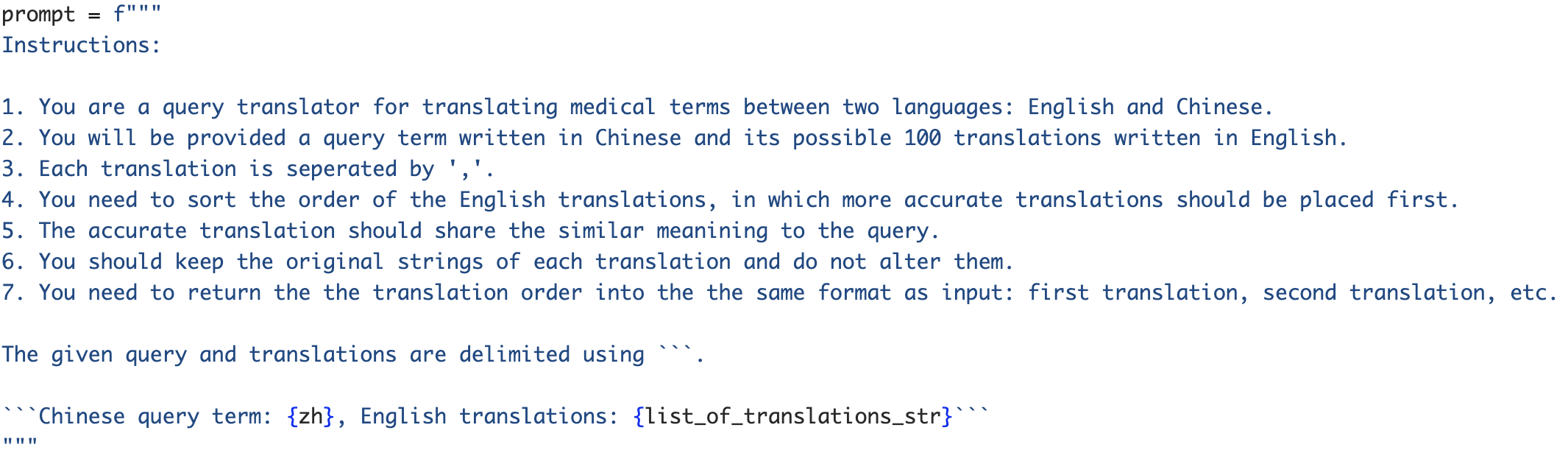}
\caption{The prompt template for asking GPT-3.5-Turbo to rerank the translations by relevance.}
\label{fig: prompt_template}
\end{figure*}

To test GPT-3.5-Turbo, we designed a prompt~(see Fig.~\ref{fig: prompt_template}) which provides GPT a query term and its list of translations. As we have annotated all 100 nearest translations for each query, we randomly shuffled all of them and asked GPT to sort the order of translations by their relevance to the query. This task was to test whether GPT could identify the semantic translation (i.e., the translation that shares the same meaning as the query) and give it a good rank (i.e., 1 is the best). The task mimicked how we used the induced bilingual health word space to search translations for each query term. To test Rerank, we called Cohere's rerank API~\footnote{\url{https://docs.cohere.com/reference/rerank-1}} and set the query and documents field as testing query and list of translations, respectively.

\begin{figure}[t]
\centering
\subfloat[$\text{ZH} \rightarrow \text{EN}$]{
\includegraphics[width=.45\textwidth]{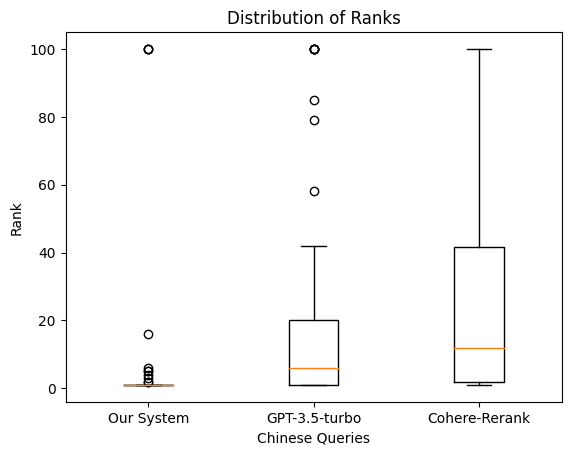} \label{subfig. Chinese query rank}}
\hfill
\subfloat[$\text{EN} \rightarrow \text{ZH}$]{
\includegraphics[width=.45\textwidth]{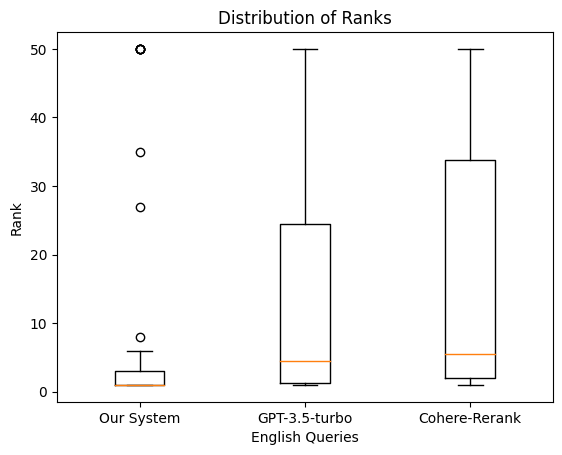}
\label{subfig. English query rank}}
\caption{The rank distributions of different systems}
\label{fig: query rank}
\end{figure}

\noindent \textbf{Results.} Table~\ref{tab:mrr-performance-comparison} reports the MRR performance for each model, in which EN $\rightarrow$ ZH means that we used English queries to find Chinese synonym candidates (i.e., Chinese translations), and ZH $\rightarrow$ EN was the reverse query direction. All three models outperformed the random baseline, indicating that all of them could suggest relevant translations for a given query. The bilingual health word embedding induced by our system reached the best MRR performance. That is, relevant translations retrieved by our system could be ranked high (supported in Fig.~\ref{fig: query rank}). To check the statistical significance, we employed the Wilcoxon signed-rank test to test the difference between the returned list of ranks across models. The results suggested that our framework significantly, $p<0.01$ marked by $\text{}^{***}$, prevailed over the other two models. Our results conformed to the finding reported in~\cite{lehmanWeStillNeed2023} that large language models may perform worse than health domain-specific models with significantly smaller parameters.
As a result, our framework was more reproducible and adopted by others because it only required relatively small corpora and a few bilingual anchors. Besides, our induced bilingual word embedding was more economically efficient, i.e., free from the need for GPU and efficient query time achieved by computing cosine similarity between two words rather than computing across layers of transformers.

\begin{table}[t]
\centering
\caption{MRR performance comparison}
\label{tab:mrr-performance-comparison}
\resizebox{\columnwidth}{!}{%
\begin{tabular}{@{}lllll@{}}
\toprule
 & Our System & GPT-3.5-Turbo & Rerank & Random Baseline \\ \midrule
ZH $\rightarrow$ EN & $0.825^{***}$ & $0.413$ & $0.322$ & $0.02$ \\
EN $\rightarrow$ ZH & $0.664^{***}$ & $0.398$ & $0.337$ & $0.04$ \\ \bottomrule
\end{tabular}%
}
\end{table}

\subsection{Experiment 2: Sensitivity Analysis}

\textbf{Performance Metrics.} We used precision, recall, and $\text{F}_1$ to report the query performance, and those metrics are defined as follows:

\begin{equation} \label{eq. Precision@N}
    \text{precision} = \frac{|\text{relevant items} \cup \text{retrieved items}|}{|\text{retrieved items}|}
\end{equation}

\begin{equation} \label{eq. Recall@N}
    \text{recall} = \frac{|\text{relevant items} \cup \text{retrieved items}|}{|\text{relevant items}|}
\end{equation}

\begin{equation} \label{eq. F1@N}
    \text{F}_1 = 2 \times \frac{\text{precision} \times \text{recall}}{\text{precision} + \text{recall}}
\end{equation}

\begin{figure}[h]%
    \centering
    \subfloat[EN $\rightarrow$ ZH]{{\includegraphics[width=\columnwidth]{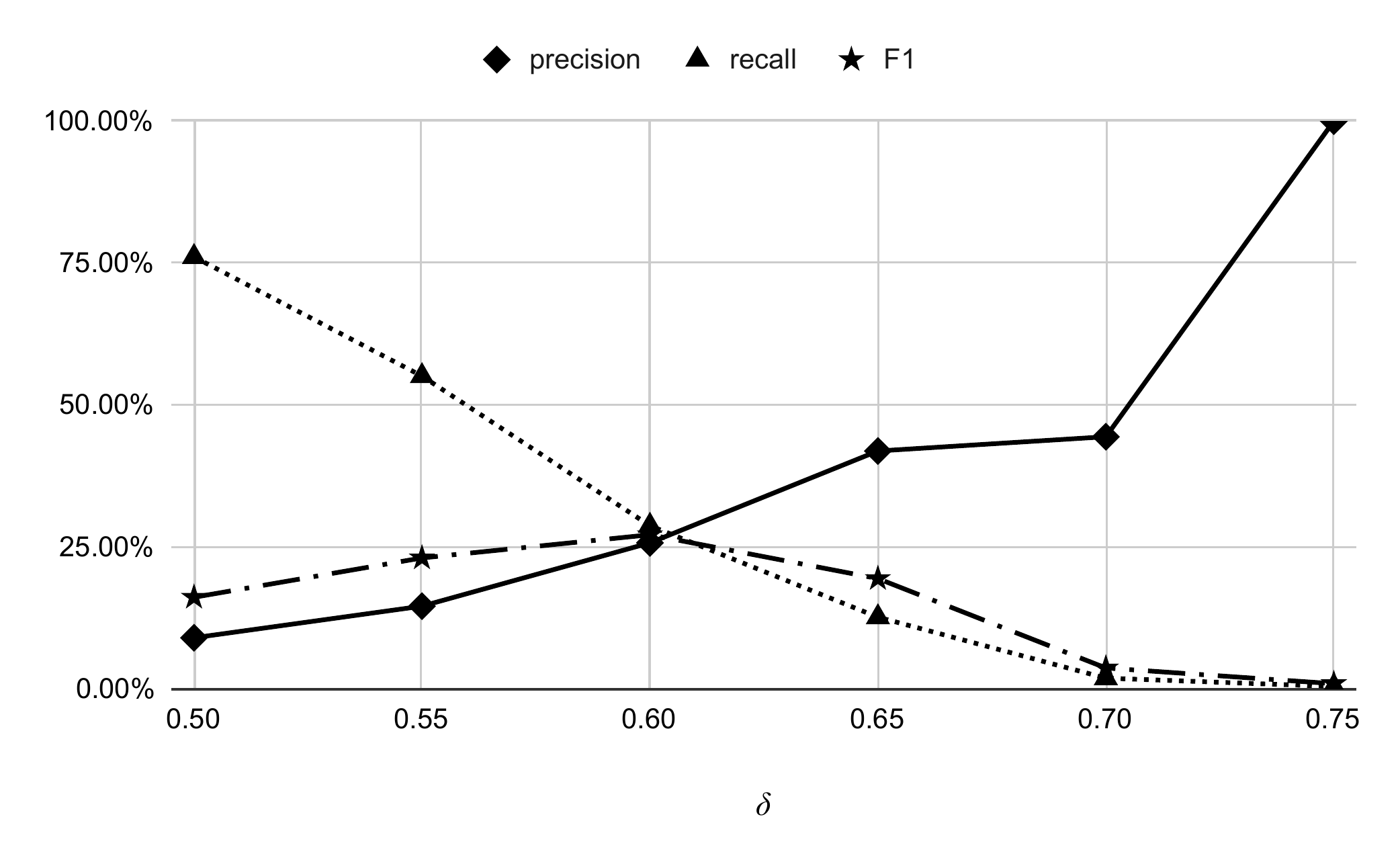}}\label{fig. exp2-en2zh-all-measures}}%
    \hfill
    \subfloat[ZH $\rightarrow$ EN]{{\includegraphics[width=\columnwidth]{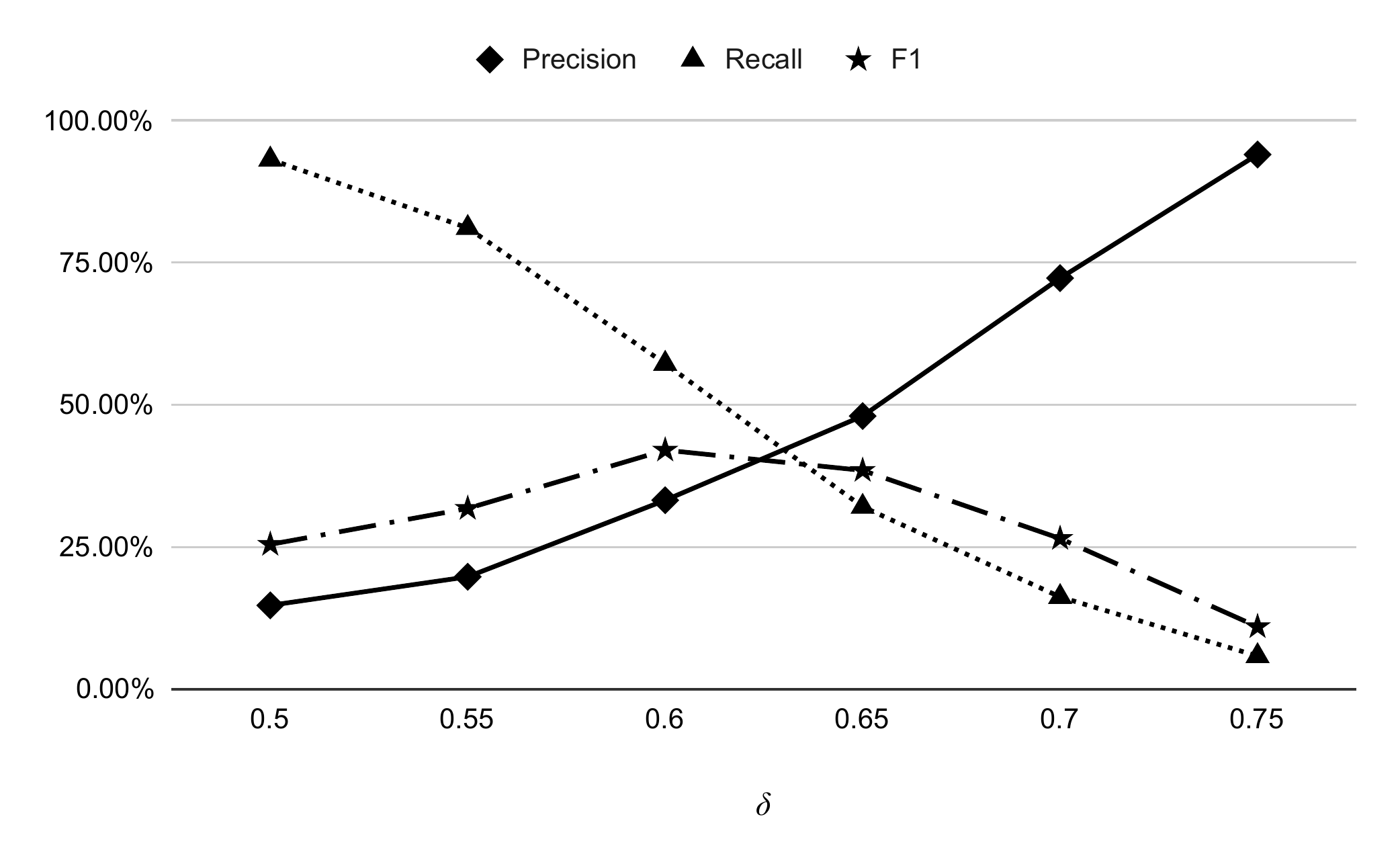}}\label{fig. exp2-zh2en-all-measures}}%
    \caption{Performance of precision, recall, and $F_1$ under different $\delta$} %
\end{figure}

\noindent \textbf{Influence on Similarity Threshold.} Fig. \ref{fig. exp2-en2zh-all-measures} and Fig. \ref{fig. exp2-zh2en-all-measures} showed the influence of a similarity threshold $\delta$ to the precision, recall, and $\text{F}_1$. Both query directions (EN $\rightarrow$ ZH and ZH $\rightarrow$ EN) had consistent trends of three metrics with respect to the x-axis, $\delta$. With the higher $\delta$, our methodology generated the more-precise synonym candidates across languages, and the precision approached the peak when $\delta = 0.75$. On the other hand, the recall was at the peak as $\delta$ stayed low; however, the price was that loose criteria included too many noisy candidates, resulting in undesirably low precision. For the task of determining bilingual medical vocabulary, both precision and recall have the same importance. Therefore, we could set 0.6 as $\delta$ for both query directions to reach the best $\text{F}_1$. Comparing with the previous studies on expanding vocabularies within the same languages \cite{Jiang2013UsingData,He2017EnrichingApproach}, our $\text{F}_1$ performance was competent even in the more challenging context---expanding vocabularies across languages. The ZH $\rightarrow$ EN had a better performance than EN $\rightarrow$ ZH. We attributed this phenomenon to the unbalanced size between corpora of different languages. Since the English language has a three times larger vocabulary than the Chinese language, it is more probable for a Chinese query to find its English synonyms.

Moreover, when we set $\delta$ at 0.6 (where $\text{F}_1$ performs the best), the correct ratio~\footnote{The correct ration is calculated by dividing number of relevant items by number of retrieved items.} of EN $\rightarrow$ ZH and ZH $\rightarrow$ EN were 25.76\%(59/229) and 33.26\%(314/944), respectively. That is, we could find one correct Chinese candidate among every 4 Chinese candidates and one correct English candidate among every 3 English candidates. This retrieval performance demonstrated that our methodology was effective for vocabulary expansion and helped to avoid scanning ten thousand possible candidates of each query. 

\begin{figure}[h]%
    \centering
    \subfloat[EN $\rightarrow$ ZH]{{\includegraphics[width=\columnwidth]{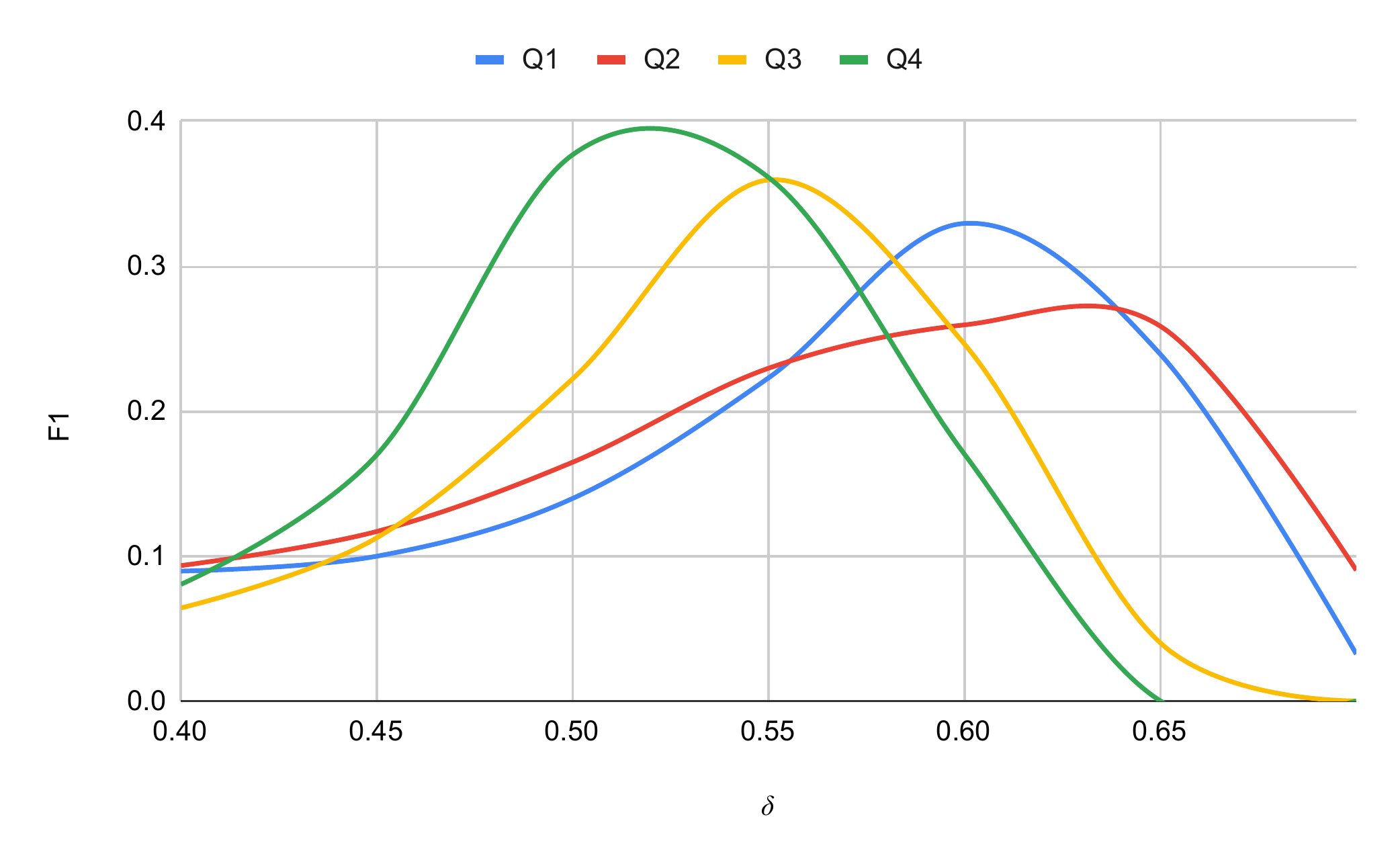}}\label{fig. exp2-en2zh-quad-f1}}%
    \hfill
    \subfloat[ZH $\rightarrow$ EN]{{\includegraphics[width=\columnwidth]{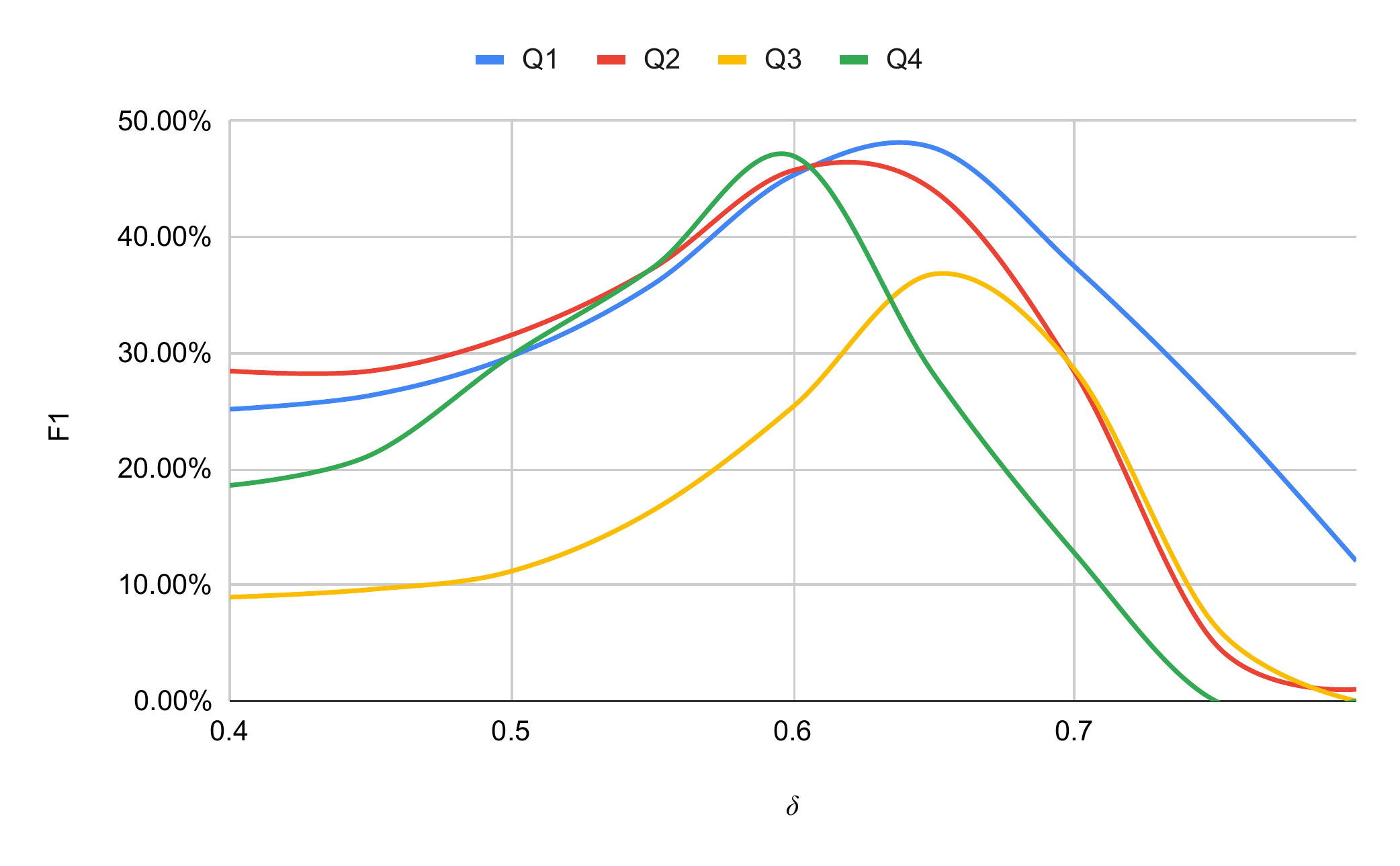}}\label{fig. exp2-zh2en-quad-f1}}%
    \caption{$F_1$ performance for different subsets of queries} %
\end{figure}

\noindent \textbf{Efficacy of Dynamic Threshold Mechanism.} To show the necessity of applying the dynamic setting of $\delta$, we ascendingly sorted the whole query set by the modularity score. We then grouped it into four quarters, namely $Q_1, Q_2, Q_3$, and $Q_4$, based on the quartile. Fig. \ref{fig. exp2-en2zh-quad-f1} and Fig. \ref{fig. exp2-zh2en-quad-f1} showed the $F_1$ curve of each quarter, in which the queries in each quarter reach the highest $F_1$ at different value of $\delta$. The curves of $Q_4$ are central on smaller $\delta$ than ones of other quarters. Instead, the curves of $Q_1$ tend to require higher $\delta$ to reach a peak of $F_1$. These patterns imply the phenomenon provided in Table \ref{tab:example of two queries with different modularity} that we cannot use a single threshold for every query. Also, the modularity score is the one potential factor to set the appropriate threshold. As a result, after grouping queries concerning modularity score, the dynamic threshold mechanism helped to adopt the threshold 
which reached the highest $F_1$ for each quarter (e.g., in EN $\rightarrow$ ZH we set $0.5$ as $\delta$ for $Q_4$ and $0.6$ for $Q_1$) to achieve better retrieval performance.

To benchmark the performance of the dynamic threshold mechanism, we compared it with the other two alternative query methods: $k$-NN and single threshold. For $k$-NN, we searched its best performance of $k$ in a range from 5 to 50 with five as step size, and for single threshold, we searched its best performance of $\delta$ in a range from 0.5 to 0.75 with 0.05 as a step size. Table \ref{tab: f1 performance of dynamic threshold} presented the performance of each method. After applying the dynamic threshold mechanism, the total candidates retrieved in ZH $\rightarrow$ EN direction shrank, yet the proportion of correct candidates increased, leading to the better $\text{F}_1$. In EN $\rightarrow$ ZH, its $\text{F}_1$ also rose, and it helped identify more correct candidates without compromising correct ratios.

\begin{table}[t]
\centering
\caption{Query performance between different methods}
\label{tab: f1 performance of dynamic threshold}
\resizebox{\columnwidth}{!}{%
\begin{tabular}{@{}lllll@{}}
\toprule
Query Direction & Query Method & \#Retrieved Items & Correct Ratio & $F_1$ \\ \midrule
\multirow{3}{*}{ZH $\rightarrow$ EN} & $k$-NN & 750 & 33.07\% & 38.21\% \\
 & Single Threshold & 944 & 33.26\% & 42.09\% \\
 & Dynamic Threshold & 716 & \textbf{40.84\%} & \textbf{44.31\%} \\ \midrule
\multirow{3}{*}{EN $\rightarrow$ ZH} & $k$-NN & 250 & \textbf{28.80\%} & 31.65\% \\
 & Single Threshold & 229 & 25.76\% & 27.19\% \\
 & Dynamic Threshold & 427 & 25.29\% & \textbf{33.13\%} \\ \bottomrule
\end{tabular}%
}
\end{table}

\subsection{Qualitative Result}

We provided the retrieved samples in Table \ref{tab: retrieved samples}. When we took 拉肚子(diarrhea) as the query, our approach helped identify not only the correct synonym--diarrhea but also some common wrong-spelling alternatives used by laypeople such as \textit{diarhea, diareah, diarreha, diarreah, diaherra, diahrrea, and diahrea}. Even though stomach\_ache and stomach\_cramp do not match the exact meaning of the query, both terms have similar contents. Those misspelled and similar contextual words may contribute to subsequent consumer-oriented health applications such as information retrieval or query expansion~\cite{changBridgingConsumerHealth2023}. On the other hand, when we set diarrhea as the query, several Chinese synonyms (i.e., 腹瀉(diarrhea), 水瀉(watery stools), 拉肚子(diarrhea), 拉水(watery stools), 瀉(diarrhea), 下痢(diarrhea), 腹泄(diarrhea), 瀉肚子(diarrhea)) were successfully identified, including two words with similar contexts: 肚痛(abdominal pain) and\text{ }消化不良(indigestion). From the above two query examples, we found that laypeople used different expressions to describe a symptom, and our framework can help identify those expressions.

\begin{table}[t]
\centering
\caption{Retrieved samples of 拉肚子(diarrhea) and diarrhea}
\label{tab: retrieved samples}
\resizebox{\columnwidth}{!}{%
\begin{tabular}{@{}llllll@{}}
\toprule
Query & Rank & Candidates & Query & Rank & Candidates \\ \midrule
\multirow{10}{*}{拉肚子(diarrhea)} & 1 & diarhea & \multirow{10}{*}{diarrhea} & 1 & 腹瀉(diarrhea) \\
 & 2 & diareah &  & 2 & 水瀉(watery stools) \\
 & 3 & diarreha &  & 3 & 拉肚子(diarrhea) \\
 & 4 & diarrhea &  & 4 & 拉水(watery stools) \\
 & 5 & diarreah &  & 5 & 瀉(diarrhea) \\
 & 6 & diaherra &  & 6 & 下痢(diarrhea) \\
 & 7 & stomach\_ache &  & 7 & 肚痛(abdominal pain) \\
 & 8 & diahrrea &  & 8 & 腹泄(diarrhea) \\
 & 9 & diahrea &  & 9 & 消化不良(indigestion) \\
 & 10 & stomach\_cramp &  & 10 & 瀉肚子(diarrhea) \\ \bottomrule
\end{tabular}%
}
\end{table}

\section{Discussion}

\subsection{Practical Implications \& Limitations}

This study presents two implications for practitioners. First, the induced non-English CHV connects to the existing English CHV thanks to the bilingual word space. Such connections help the induced non-English medical terms conform to the existing medical terminology, such as concept unique identifiers (CUI) in UMLS. Therefore, it is possible to construct a language-agnostic CHV for benefiting multilingual applications (e.g., analyzing code-switching medical texts). Second, since the cross-lingual ATR learns word association patterns from the collected HCGC, it helps capture the commonly used medical expressions from laypeople, facilitating subsequent consumer-oriented applications such as medical chatbots. The learned vocabulary also helps translate professional medical terms into consumer-friendly terms~\cite{moramarcoMorePatientFriendly2022a} (i.e., language simplification task), improving the communication between patients and physicians. 

There are two limitations to our proposed method. The first one is that the technique we adopt for handling multiword expressions may not be effective for infrequent expressions. We employ phrase identification~\cite{Mikolov2013DistributedCompositionality} to capture the frequent term co-occurrences as multiword expressions, such as ``loose bowel movements'' and ``blood pressure problem''. Therefore, some of the less frequent multiword expressions may be ignored. The second one is that the unbalanced size of corpora between languages could harm synonym retrieval performance. For example, in our evaluation, the size of the Chinese corpus is significantly less than one of the English corpus, so we find the query direction EN $\rightarrow$ ZH has worse performance than that of ZH $\rightarrow$ EN because of the limited potential synonym candidates in Chinese.

\subsection{Future Directions}

Our proposed method can be generalized to align word vector spaces of two more languages to compare consumer-oriented expressions across multiple languages. The framework can be used for exploring different vocabulary usage patterns in different countries and generate a language-agnostic CHV for facilitating cross-lingual medical information retrieval. Another direction is to develop a query suggestion tool using our cross-lingual ATR framework. Using the appropriate terms in the query is critical to consumer health search \cite{Teixeira_Lopes2017-kt,changBridgingConsumerHealth2023}. Since every non-English native consumer may have different English literacy, recommending common-used medical expressions as a query is potentially helpful to most consumers for having results with better readability \cite{TeixeiraLopes2019InterplayTerminology}. Lastly, our framework may be extended to consider more guidances from the pre-defined English medical terminology. Medical professionals have constructed several medical terminologies, such as MedDRA, in a hierarchical structure defined by the relationships between medical entities. Once that English human-crafted knowledge can be propagated across different languages, it will benefit most other languages.

\section{Conclusion}

To automatically expand English OAC CHV into other languages, we proposed a cross-lingual ATR framework, capturing word semantics from HCGC within each language using the word embedding technique. The words of different languages will be aligned into a cross-lingual word space by referring to a few pairs of medical word translations. The resultant cross-lingual word space effectively retrieves similar health-related expressions across languages, such as misspelling words and colloquial terms. Our framework prevents vocabulary organizers from screening thousands of candidates to select appropriate translations.

\end{CJK*}

\ifCLASSOPTIONcompsoc
  \section*{Acknowledgments}
\else
  \section*{Acknowledgment}
\fi

This work was supported in part by the National Science Foundation under the Grants IIS-1741306 and IIS-2235548, and by the Department of Defense under the Grant DoD W91XWH-05-1-023.  This material is based upon work supported by (while serving at) the National Science Foundation.  Any opinions, findings, and conclusions or recommendations expressed in this material are those of the author(s) and do not necessarily reflect the views of the National Science Foundation.

\bibliographystyle{IEEEtran}
\bibliography{main}

\end{document}